\documentclass{article}

\usepackage{preamble}

\begin{document}
    
\title{Bayesian topological convolutional neural nets}

%\date{September 9, 1985}	% Here you can change the date presented in the paper title
% \date{} 					% Or remove it

\author{Sarah Harkins~Dayton$^*$ \\
	Department of Mathematics \\
	The University of Tennessee, Knoxville \\
	Knoxville, TN 37996 \\
	\texttt{sharkin5@vols.utk.edu} \\
	%% examples of more authors
	\And
	Hayden~Everett\thanks{Equal contribution} \\
	Department of Mathematics \\
	The University of Tennessee, Knoxville\\
	Knoxville, TN 37996\\
	\texttt{heveret1@vols.utk.edu} \\
	\AND
	Ioannis Schizas \\
	DEVCOM ARL \\
    Army Research Lab \\
	Aberdeen, MD 21001 \\
	\texttt{ioannis.d.schizas.civ@army.mil} \\
	\And
	David L. Boothe Jr. \\
	DEVCOM ARL \\
    Army Research Lab \\
	Aberdeen, MD 21001 \\
	\texttt{david.l.boothe7.civ@army.mil} \\
	\And
	Vasileios Maroulas \\
	Department of Mathematics \\
    The University of Tennessee, Knoxville \\
	Knoxville, TN \\
	\texttt{vmaroula@utk.edu} \\
}

% Uncomment to remove the date
\date{}

% Uncomment to override  the `A preprint' in the header
\renewcommand{\headeright}{}
\renewcommand{\undertitle}{}
\renewcommand{\shorttitle}{}

%%% Add PDF metadata to help others organize their library
%%% Once the PDF is generated, you can check the metadata with
%%% $ pdfinfo template.pdf
\hypersetup{
pdftitle={Bayesian topological convolutional neural nets},
% pdfsubject={q-bio.NC, q-bio.QM},
pdfauthor={Sarah Harkins~Dayton, Hayden~Everett, Ioannis~Schizas, David L.~Boothe Jr., Vasileios~Maroulas},
pdfkeywords={Bayesian neural network, convolutional neural network, model calibration, topological deep learning, uncertainty quantification}
}

\maketitle

\begin{abstract}
	Convolutional neural networks (CNNs) have been established as the main workhorse in image data processing; nonetheless, they require large amounts of data to train, often produce overconfident predictions, and frequently lack the ability to quantify the uncertainty of their predictions. To address these concerns, we propose a new Bayesian topological CNN that promotes a novel interplay between topology-aware learning and Bayesian sampling. Specifically, it utilizes information from important manifolds to accelerate training while reducing calibration error by placing prior distributions on network parameters and properly learning appropriate posteriors. One important contribution of our work is the inclusion of a consistency condition in the learning cost, which can effectively modify the prior distributions to improve the performance of our novel network architecture. We evaluate the model on benchmark image classification datasets and demonstrate its superiority over conventional CNNs, Bayesian neural networks (BNNs), and topological CNNs. In particular, we supply evidence that our method provides an advantage in situations where training data is limited or corrupted. Furthermore, we show that the new model allows for better uncertainty quantification than standard BNNs since it can more readily identify examples of out-of-distribution data on which it has not been trained. Our results highlight the potential of our novel hybrid approach for more efficient and robust image classification.
\end{abstract}

% keywords can be removed

%%%%%%%%%%%%%%%%%%%% Introduction %%%%%%%%%%%%%%%%%%%%%%%%%%
%\normalsizefont

\section{Introduction}
Convolutional neural networks (CNNs) have become one of the most widely used tools in machine learning for image processing. Although they can attain high accuracy on certain image classification tasks, standard CNNs often require large amounts of training data to achieve good performance and are prone to overfitting on small datasets \cite{implicit_regularization}. Moreover, current state-of-the-art networks have many hidden layers, which can lead to poor model calibration and overconfident predictions \cite{onCalibration}.

Topological data analysis (TDA) refers to a broad array of techniques that seek to discover and utilize the geometric shape of data to extract information from it, a feature which is not present in traditional CNNs \cite{hajij2023topologicaldeeplearninggoing, papamarkou2024positiontopologicaldeeplearning, TDAinBioMedicine}. Nevertheless, TDA has been employed to show that the convolutional filter weights of trained CNNs correspond to points on the Klein bottle \cite{TAtDL_Carlsson}, a finding that is in line with previous results demonstrating that patches of natural images tend to cluster around an embedding of the Klein bottle \cite{Carlsson_NaturalImages}. Motivated by these empirical observations, Love et al. \cite{LoveTCNN} introduced the topological convolutional neural network (TCNN), which applies topological methods to modify the convolutional layers. Specifically, discretizations of manifolds are used either to fix the weights used in convolutional filters or to prune connections between neurons in successive convolutional layers during training. The authors provide evidence that TCNNs are more accurate and generalizable compared to standard CNNs and can learn more quickly. However, their work does not address model calibration in TCNNs, which can be important to users interested in knowing when a network's predictions can be trusted and when greater skepticism may be warranted. We have also found that the TCNN model struggles in data starvation situations and when the training set is contaminated by random perturbations.

Compared to standard architectures that learn point estimates of the weights in a neural network, Bayesian neural networks (BNNs) have been shown to be less prone to overfitting, especially on small datasets, and better calibrated \cite{hands_on_BNN, primer_on_BNNs}. Moreover, Bayesian techniques have been employed to increase the predictive accuracy of CNNs \cite{gal2016bayesianconvolutionalneuralnetworks} and improve their robustness to adversarial attacks \cite{BayesianGCNN}. In spite of these advantages, BNNs are less commonly used since training them is usually much more computationally expensive \cite{papamarkou2024positionbayesiandeeplearning}. Increasing the scalability of BNNs and Bayesian optimization is an active area of research \cite{ScalableBNNs_rank1, moslemi2024scalingbayesianneuralnetworks, Scalable_BO}.

To remedy challenges that arise with model calibration and low-data settings, we propose a Bayesian topological CNN (BTCNN). The BTCNN introduces stochasticity in the network parameters by considering pertinent prior distributions and carefully learning the corresponding posterior distributions, while simultaneously accelerating the training process through the inclusion of topological convolutional layers. Our novel method takes advantage of topological features of the data and yields a network that achieves high accuracy and produces more appropriately confident predictions even when training data is sparse or noisy. Furthermore, the BTCNN can be transferred to different scenarios by incorporating a consistency condition into the prior distributions to enhance model calibration. To the best of our knowledge, this is the first attempt that explicitly incorporates topological convolutional layers and Bayesian inference into a unified neural network architecture for improved data classification. Our main contributions are the following:
\begin{enumerate}
    \item Topological and Bayesian components are integrated into a CNN architecture to achieve better accuracy and calibration on image classification tasks.
    \item A prior distribution on the network weights is introduced using a consistency condition, and a new loss function is considered.
    \item We provide evidence that our method outperforms standard CNNs, TCNNs, and BNNs, especially in cases where data sets are small or noisy.
    \item We show that our model leads to superior uncertainty quantification compared to conventional BNNs, as it produces more uncertain predictions for out-of-distribution data on which it has not been trained and lower uncertainty on in-distribution samples.
\end{enumerate}
The paper is organized as follows. Section \ref{sec:Background} summarizes relevant background information. Next, we provide a detailed description of the BTCNN architecture and use a consistency condition to formulate an adjusted loss function to train the network in Section \ref{sec:Methodology}. In Section \ref{sec:Results}, we test the BTCNN's performance with and without the consistency condition on various image classification tasks and compare our model to standard CNNs, TCNNs, and BNNs. Finally, we conclude in Section \ref{sec:Discussion} with a discussion of our findings, limitations of our method, and potential future investigations. 

%%%%%%%%%%%%%%%%%%%% Background %%%%%%%%%%%%%%%%%%%%%%%%%%
\section{Background}\label{sec:Background}

\subsection{Problem Statement}
Consider a supervised image classification problem where $\textbf{X} = \{\mathbf{x}_1, \mathbf{x}_2, \dots , \mathbf{x}_N \}$ are the training set inputs and $\textbf{Y} = \{y_1, y_2, \dots , y_N \}$ are the corresponding labels. The goal is to approximate the true relationship $y=f(\mathbf{x})$ between an input $y$ and an output $\mathbf{x}$ by learning the parameters $\theta$ of a neural network $\Phi$. This approximation 
\begin{equation}\label{function approximation}
    \Phi(\mathbf{x};\theta)\approx f(\mathbf{x})
\end{equation}
is ideally able to accurately classify the elements of $\mathbf{X}$ while also generalizing to new examples not included in the training set. Challenges arise when there is only a limited amount of training data or if the training images are low-quality, distorted, or perturbed in some way (e.g., corrupted by random noise, blurring, or incorrect labels). Some preliminary background information is presented next.

\subsection{Bayesian Neural Networks}\label{bnn background}
When using deterministic neural networks, one attempts to learn the best point estimate $\hat{\theta}$ for the parameters of the network $\Phi$, as defined in equation (\ref{function approximation}). In contrast, BNNs operate by placing a prior distribution $p(\theta)$ on the network parameters $\theta$ and then learning the posterior distribution $p(\theta|\mathcal{D})$ of the parameters given the training data $\mathcal{D}=\{\mathbf{X},\mathbf{Y}\}$ using Bayes' rule and that fact that inputs are independent of the model parameters:
\begin{equation}\label{posterior}
    p(\theta|\mathcal{D}) = \frac{p(\textbf{Y}| \textbf{X}, \theta) p(\theta)}{\int_{\theta} p(\textbf{Y}| \textbf{X}, \theta') p(\theta') \,d\theta'} \propto p(\textbf{Y}| \textbf{X}, \theta) p(\theta).
\end{equation}
Provided that Equation (\ref{posterior}) can be computed or approximated, the predictive distribution for the label $y$ corresponding to a new input $\mathbf{x}$ given the observations $\mathcal{D}$ is computed as 
\begin{equation}\label{predictive distribution}
    p(y|\mathbf{x},\mathcal{D}) = \int_\theta p(y|\mathbf{x}, \theta') p(\theta'| \mathcal{D}) \,d\theta'.
\end{equation}
The function $p(y|\mathbf{x},\theta)$ represents the probability of assigning the label $y$ to an input $\mathbf{x}$ given a particular set of parameters $\theta$ and must be specified in advance. For example, one can choose to model $p(y|\mathbf{x},\theta)$ with a categorical distribution when performing classification. In practice, the integral in Equation (\ref{predictive distribution}) is intractable, but this problem can be avoided by repeatedly sampling from the posterior and then performing the approximation
\begin{equation}\label{approx bma}
    p(y|\mathbf{x},\mathcal{D})\approx\hat{\mathbf{p}} = \frac{1}{S}\sum_{s=1}^{S} \Phi\left(\mathbf{x};\theta^{(s)}\right),
\end{equation}
where $S$ instances of the parameters $\theta^{(s)}$ are drawn from $p(\theta|\mathcal{D})$. For classification, observe that $\hat{\mathbf{p}}$ is a vector whose elements are the predicted probabilities that $\mathbf{x}$ belongs to each of the different classes. The output classification is then provided by $\hat{y} = \argmax_{j} \hat{p}_j \in \hat{\mathbf{p}}$, that is, by selecting the class corresponding to the index of the largest entry of $\hat{\mathbf{p}}$.

Given that $p(\theta|\mathcal{D})$ frequently cannot be represented in closed form, several classes of methods have been developed to estimate the posterior distribution. Markov Chain Monte Carlo (MCMC) methods draw samples directly from the posterior; however, these techniques are often computationally intensive even with relatively small datasets \cite{hands_on_BNN}. Variational inference (VI) is a more scalable alternative to MCMC that approximates the true posterior $p(\theta|\mathcal{D})$ by learning the parameters $\phi$ of a simpler distribution $q_\phi(\theta)$ so that the two are as close as possible. Closeness is measured using the Kullback-Leibler (KL) divergence, denoted $D_{KL}$, so VI amounts to finding
\begin{equation}\label{phi mle}
    \hat{\phi} = \argmin_{\phi}D_{KL}[q_\phi(\theta)||p(\theta|\mathcal{D})],
\end{equation}
where the KL divergence between the true and variational posteriors is calculated via 
\begin{equation*}
    D_{KL}[q_\phi(\theta)||p(\theta|\mathcal{D})]=\int_\theta q_\phi(\theta')\log\left[\frac{q_\phi(\theta')}{p(\theta'|\mathcal{D})}\right]\,d\theta'.
\end{equation*}
Equation (\ref{phi mle}) can be equivalently expressed as
\begin{equation}\label{variational free energy}
    \hat{\phi} = \argmin_{\phi} D_{KL}[q_\phi(\theta)||p(\theta)]-\int_\theta q_\phi(\theta')\log p(\mathbf{Y}|\mathbf{X},\theta')\,d\theta',
\end{equation}
which is easier to work with since it does not require knowledge of the exact form of the true posterior. The above formulation of VI as a minimization problem provides a cost function that is appropriate for acquiring the best estimates of $\phi$ using the gradient descent approach common in machine learning. Once the optimal parameter estimates are obtained, predictions are made using Equation (\ref{approx bma}) by sampling network parameters from $q_\phi(\theta)$ \cite{primer_on_BNNs}.

When training a traditional neural network, the usual procedure is to find estimates $\hat\theta$ for the network parameters that minimize a predetermined cost function, which is often the sum of a data-dependent loss term $\mathcal{L}(\mathcal{D},\theta)$ and a data-independent regularization term $\mathcal{R}(\theta)$ whose goal is to improve the generalizability of the learned model. In other words, one seeks to find $\hat{\theta}=\argmin_{\theta}\mathcal{L}(\mathcal{D},\theta)+\mathcal{R}(\theta)$. For BNNs, the goal is to find parameter estimates $\hat{\theta}$ that maximize the posterior $p(\theta|\mathcal{D})$, so by letting $\mathcal{L}(\mathcal{D},\theta)=-\log p(\mathbf{Y}|\mathbf{X},\theta)$ and $\mathcal{R}(\theta)=-\log p(\theta)$, it follows that the choice of prior in a BNN is analogous to the choice of a regularization term in a traditional neural network.

When training a BNN, it may be beneficial to use a consistency condition $C(\theta, \mathbf{x})$ that quantifies how well the network's predictions obey certain criteria. Introducing a consistency condition can be thought of as a form of regularization where $C(\theta, \mathbf{x})$ is the log-likelihood of a prediction given network parameters $\theta$ and input $\mathbf{x}$ \cite{hands_on_BNN}. Accordingly, the consistency condition may be included with the prior so that, in effect, the prior is conditioned on the inputs $\mathbf{X}$. In this case, the distribution for the prior with the consistency condition satisfies
\begin{equation}\label{prior with cc}
    p(\theta|\mathbf{X})\propto p(\theta)\exp\left\{-\frac{1}{|\mathbf{X}|}\sum_{\mathbf{x}\in\mathbf{X}}C(\theta,\mathbf{x})\right\}.
\end{equation}
This modification of the prior distribution has the effect of adding an additional term to the cost function used to train a BNN. Later on we will consider a specific  functional form to encourage the network to provide similar outputs for data belonging to the same class.

\subsection{Uncertainty Quantification}
When considering a learning model, there are two major types of uncertainty: epistemic and aleatoric. Epistemic uncertainty captures uncertainty in the model itself, i.e., whether the model parameters are poorly determined due to insufficient data. Aleatoric uncertainty accounts for uncertainty in the observations \cite{gawlikowski2023}. One key difference between epistemic and aleatoric uncertainty is that epistemic uncertainty can be reduced by introducing more data while aleatoric uncertainty cannot (assuming the measurement precision of the data is held constant) \cite{Gal2016Uncertainty}. One metric for measuring uncertainty is the entropy of the predictive distribution $\mathcal{H}(\mathbf{\hat{p}}) = -\sum_{c=1}^C \hat{p}_{c}\cdot\log(\hat{p}_{c}),$ where $\hat{\mathbf{p}}$ is the probability vector and $C$ is the number of classes \cite{kendall2017uncertaintiesneedbayesiandeep}. However, this method does not differentiate between epistemic and aleatoric uncertainty \cite{smith2018understandingmeasuresuncertaintyadversarial}.

To distinguish between the two, we compute the difference between the total and aleatoric uncertainties by utilizing the predictive distribution, entropy, and mutual information as described in \cite{depeweg2018decompositionuncertaintybayesiandeep, distelzweig2024entropybaseduncertaintymodelingtrajectory, he2025surveyuncertaintyquantificationmethods, smith2018understandingmeasuresuncertaintyadversarial}  
\begin{equation}\label{uncertainty breakdown}
    \mathcal{I}(y, \theta | \mathbf{x},\mathcal{D}) = \mathcal{H}[p(y| \mathbf{x},\mathcal{D} )] - \mathbb{E}_{p(\theta | \mathcal{D})} (\mathcal{H}[p(y|\mathbf{x}, \theta)]).
\end{equation}

In Equation (\ref{uncertainty breakdown}), $\mathcal{I}(y, \theta|\mathbf{x},\mathcal{D})$ represents the mutual information between the model's prediction, $y$, and the model parameters, $\theta$. Thus, the mutual information measures the epistemic uncertainty of the model. The term $\mathcal{H}[p(y| \mathbf{x},\mathcal{D})]$ represents the total uncertainty of the predictive distribution, called the predictive entropy. The term $\mathbb{E}_{p(\theta | \mathcal{D})} (\mathcal{H}[p(y|\mathbf{x}, \theta)])$ represents the expected entropy, or aleatoric uncertainty.

Because BNNs have stochastic weights, at the time of inference, the network weights are sampled indirectly from the posterior, and model averaging is done to compute the relative probability of each class. This sampling of model weights simulates having multiple possible models. For this reason, BNNs are considered a special case of ensemble learning \cite{hands_on_BNN}. Let $F = \left\{\Phi_{\theta^{(s)}}\right\}^S_{s=1}$ be the set of Bayesian models with the corresponding set of parameters $\Theta = \left\{\theta^{(s)}\right\}^S_{s=1}.$ The set $F$ defines our ensemble containing $S$ ensemble members, $\Phi_{\theta^{(s)}}$. The model prediction will be calculated as described in Equation (\ref{approx bma}), i.e., via model averaging. In practice, the true posterior is not available, thus we will replace $p(\theta|\mathcal{D})$ with $q_\phi(\theta)$. Utilizing MC approximation and the definition of entropy,
\begin{equation}\label{uncert approx}
    \mathcal{I}(y, \theta|\mathbf{x},\mathcal{D}) \approx -\sum_{c=1}^C \hat{p}_c \cdot \log_2(\hat{p}_c) - \frac{1}{S} \sum_{s=1}^S  \mathcal{H} \left[\Phi\left(\mathbf{x};\theta^{(s)}\right)\right]
\end{equation}
where 
\begin{equation}
    \mathcal{H} \left[\Phi\left(\mathbf{x};\theta^{(s)}\right)\right] = -\sum_{c=1}^{C} \left[\Phi\left(\mathbf{x};\theta^{(s)}\right)\right]_c \log_2\left[\Phi\left(\mathbf{x};\theta^{(s)}\right)\right]_c
\end{equation}
is the entropy of ensemble member $\Phi_{\theta^{(s)}}$ \cite{Olmin_2022}. 

In general, a uniform distribution maximizes entropy. Thus, total uncertainty is high when the model prediction $\hat{\mathbf{p}}$ is flat, i.e., when there is a lot of uncertainty in $\hat{\mathbf{p}}$. The probability vector $\hat{\mathbf{p}}$ can be flat for a few reasons. One reason for high total uncertainty is disagreement among the ensemble members. Ensemble member disagreement occurs when the member predictions are rather certain, but the class of high probability differs among the members. When there is disagreement among the ensemble members, the ensemble prediction will be flattened due to model averaging. In this case, the aleatoric uncertainty will be low, thus causing the epistemic uncertainty to be high. Another reason for high total entropy is that each of the ensemble members has a relatively high entropy, that is, each ensemble member has a relatively flat distribution. In this case, the aleatoric uncertainty will be high, causing the epistemic uncertainty to be low. 
%%%%%%%%%%%%%%%%%%%% Methodology %%%%%%%%%%%%%%%%%%%%%%%%%%
\section{Bayesian Topological Convolutional Neural Network}\label{sec:Methodology}
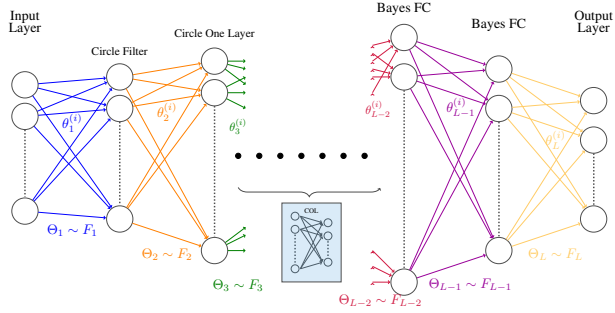
\begin{figure}[!t]
    \centering
    \scalebox{.35}{\begin{tikzpicture}[
    scale=1.2,
    shorten >=1pt,->,draw=black!70, 
    node distance=\layersep,
    neuron/.style={circle,fill=black!25,minimum size=20,inner sep=0},
    edge/.style 2 args={pos={(mod(#1+#2,2)+1)*0.33}, font=\tiny},
    distro/.style 2 args={
        edge={#1}{#2}, node contents={}, minimum size=1cm, path picture={\draw[double=TennesseeOrange,white,ultra thick,double distance=1pt,shorten >=0pt] plot[variable=\t,domain=-1:1,samples=80] ({\t},{0.25*exp(-100*(\t-0.25*(#1-2.5))^2 ))});}
      },
    weight/.style 2 args={
        edge={#1}{#2}, node contents={\pgfmathparse{0.35*#1-#2*0.15}\pgfmathprintnumber[fixed]{\pgfmathresult}}, fill=white, inner sep=2pt
      },
    fill opacity = 0.2, % opacity of filled-in rectangles 
    truck2/.style={
        minimum width={4cm+0.4pt},
        minimum height={6cm+0.4pt},
        path picture={
            \pic at ([yshift=0.2pt]path picture bounding box.center) {truck2};
        },
        yshift=0.1cm
        },
  ]
%({\t},{0.25*exp(-100*(\t-0.05*(#1-2.5))^2 - 3*\t*#2))});}

  \begin{scope}[xshift=10cm]
    \genericarchcolor{bayes}
  \end{scope}

  % Draw distros for all Bayesian edges.
  % \foreach \i in {1,...,3}
  %   \foreach \j in {1,...,3}
  %       \path (fc1\i-bayes) -- (fc2\j-bayes) node[distro={\i}{\j}];
  % \foreach \i in {1,...,3}
  %     \foreach \j in {1,...,3}
  %       \path (fc2\i-bayes) -- (output\j-bayes) node[distro={\i}{\j}];
        
\end{tikzpicture}}
    \caption{Overview of the generic BTCNN architecture. For each layer $k \in \{1, \dots , L \}$, the weight of connection $i$ in layer $k$ of the model follows the distribution $F_{k}$, that is $\theta_{k}^{(i)} \sim F_{k}$. }
    \label{fig:btcnn_layers}
\end{figure}
The BTCNN is a hybrid neural architecture that unifies topological inductive biases and Bayesian inference to improve generalization performance and uncertainty quantification which leads to more reliable responses when the network is presented with data not conforming to the training distribution. The BTCNN is composed of two major components: (1) a topological feature extractor, which uses discretized manifolds to model structural invariances in data, and (2) a Bayesian classifier, which introduces probability distributions over parameters to capture uncertainty in predictions. The overall architecture processes input data through topological convolutional layers before passing the results through fully connected (dense) layers. Prior distributions can be imposed on the topological convolutional layers and/or the fully connected layers as decided by the user. This design allows our method to retain the benefits of structural priors while enabling principled uncertainty estimates in its predictions.

The BTCNN differs from standard BNNs in that it explicitly incorporates topological constraints into the model's structure. In contrast to conventional BNNs, which treat all features as equally distributed and independent, our approach encodes invariances through symmetries in the primary circle ingrained in its filter design. Conversely, unlike TCNNs that lack uncertainty quantification, the BTCNN allows for probabilistic reasoning and predictive calibration. By combining both topological inductive biases and probability distributions (Bayesian inference), the network offers a structurally regularized, uncertainty-aware model suitable for domains where spatial consistency and reliability are critical.

The front end of our network is designed to encode topological structures resulting from prior research on natural image data \cite{Carlsson_NaturalImages} via topological filters, specifically the \textit{Circle Filter (CF)} and the \textit{Circle-One Layer (COL)} introduced in \cite{LoveTCNN}. The CF layer is constructed using a fixed set of filters defined over the unit circle \( S^1 \subset \mathbb{R}^2 \). The CF layer is a convolutional layer where the filter is initialized on an embedding of the primary circle, $F_{S^1}(x)(t,u) = \cos(x)t + \sin(x)u$, where $S^1 : = \{\kappa \in \mathbb{R}^2 : |\kappa| = 1 \}$. The filter weights are fixed, that is, they do not change during training. The COL layer extends this idea by constraining network parameters to follow a circular topology. Rather than learning unconstrained filters in \( \mathbb{R}^2 \), weights are initialized and optimized along a circular parameterization, effectively learning localized representations that preserve topological consistency. The COL identifies each filter in a convolutional layer with a point on the primary circle and then, given a specified metric and threshold, zeroes out the weights between filters whose distance from each other on the primary circle exceeds the threshold.

\begin{figure}[!t]
    \centering
    \includegraphics[width=0.5\linewidth]{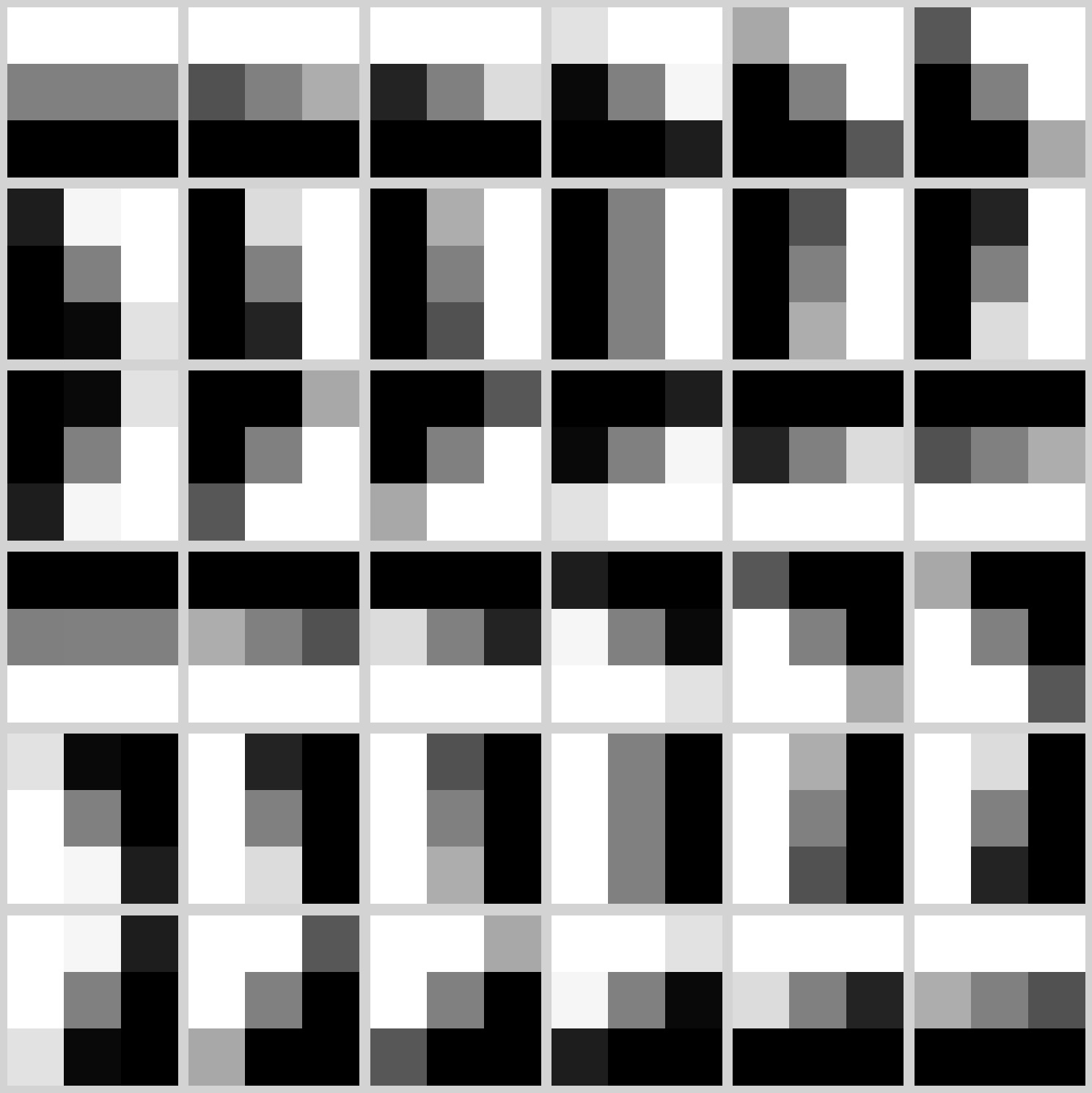}
    \caption{An array of 36 3x3 circle filters.}
    \label{fig:circle_filter_init}
\end{figure}

Stochasticity can be introduced to all trainable network parameters; however, making a layer Bayesian by placing a prior distribution $p(\theta)$ on a layer's weights and biases and learning the appropriate posterior distribution $p(\theta|\mathcal{D})$ using VI generally entails increasing the number of parameters compared to a non-Bayesian layer (e.g., for every connection between two neurons, one must learn a mean $\mu$ and standard deviation $\sigma$ instead of a scalar weight $w$ when using a Gaussian prior). Our novel learning architecture offers a trade-off between computational complexity during training and the flexibility offered by the Bayesian layers.

We now show how to obtain the approximate cost function used to train the BTCNN with a consistency condition included in the prior. During training, $p(\theta|\mathcal{D})$ is assumed to belong to a family of variational distributions $q_\phi(\theta)$ parametrized by $\phi$. To ensure that Equation (\ref{variational free energy}) is satisfied with $p(\theta)$ replaced with $p(\theta|\mathbf{X})$ from Equation (\ref{prior with cc}), we define the cost function $\mathcal{F}(\mathcal{D},\phi)$ as
\begin{equation}\label{bbb cost}
    \mathcal{F}(\mathcal{D},\phi) 
    =D_{KL}[q_\phi(\theta)||p(\theta|\mathbf{X})]-\int_\theta q_\phi(\theta')\log p(\mathbf{Y}|\mathbf{X},\theta')\,d\theta'.
\end{equation}
To efficiently learn $\phi$ using gradient-based techniques, we mirror the approach proposed in \cite{blundell2015weightuncertaintyneuralnetworks} by re-expressing the model parameters $\theta$ as a deterministic function of the variational parameters $\phi$ and random noise $\epsilon\sim p(\epsilon)$. Specifically, we initialize $\phi$, sample a random draw of $\epsilon$ from $p(\epsilon)$, and then apply a deterministic function $g$ to $\epsilon$ and $\phi$ in such a way that $\theta=g(\epsilon,\phi)$ is distributed according to $q_\phi(\theta)$. For example, if $q_\phi(\theta;)$ is assumed to be a multivariate normal distribution with mean $\mu$ and diagonal covariance matrix $\Sigma$, one can let $\phi=(\mu,\sigma)$ where $\sigma$ is a vector of the diagonal elements of $\Sigma^{1/2}$, sample $\epsilon$ from $\mathcal{N}(0,I)$ and apply the function $g(\epsilon,\phi)=\mu+\epsilon\odot\sigma$. This expression of $\theta$ as a deterministic function of $\epsilon$ and $\phi$ is done to enable the calculation of the derivative of the cost function with respect to $\phi$, which cannot be done when $\theta$ is a random variable. We then use Monte-Carlo samples to empirically approximate Equation (\ref{bbb cost}) as 
\begin{equation}\label{approx bbb loss}
        \hat{\mathcal{F}}(\mathcal{D},\phi)=\frac{1}{T}\sum_{t=1}^T\Bigg\{\log\left[\frac{q_\phi\left(\theta^{(t)}\right)}{ p\left(\theta^{(t)}\right)p\left(\mathbf{Y}|\mathbf{X},\theta^{(t)}\right)}\right] + \frac{1}{|\mathbf{X}|}\sum_{\mathbf{x}\in\mathbf{X}}C\left(\theta^{(t)},\mathbf{x}\right)\Bigg\}
\end{equation}
where the samples $\epsilon^{(t)}$, $t=1,...,T$ are drawn from $p(\epsilon)$, yielding the realizations $\theta^{(t)}=g\left(\epsilon^{(t)},\phi\right)$, $t=1,...,T$. When imposing the consistency condition on the BTCNN, we let
\begin{equation}\label{consistency condition}
    C(\theta,\mathbf{x})=\gamma\sum_{\tilde{\mathbf{x}}\neq\mathbf{x}}\frac{||\Phi(\mathbf{x};\theta)-\Phi(\tilde{\mathbf{x}};\theta)||_2^2}{||\mathbf{x}-\tilde{\mathbf{x}}||_F^2},
\end{equation}
where $||\cdot||_F$ denotes the Frobenius norm and $\gamma$ is a tunable hyperparameter that controls how much importance should be given to the consistency term in the cost function. Intuitively, we expect the network to produce similar outputs for similar input images, and the form of our consistency condition is designed to promote this behavior. When the difference $||\mathbf{x}-\tilde{\mathbf{x}}||_F$ is small, the network will be penalized by the addition of a large term to the cost function unless $||\Phi(\mathbf{x};\theta)-\Phi(\tilde{\mathbf{x}};\theta)||_2$ is also small. The consistency condition thus encourages the network to assign comparable probability distributions to images that are similar, i.e., their difference has a small Frobenius norm. When $\gamma=0$, Equation (\ref{approx bbb loss}) reduces to the approximate cost function for a BTCNN without a consistency condition (in other words, it approximates a true cost similar to the one given in Equation (\ref{bbb cost}) but with $p(\theta|\mathbf{X})$ replaced with $p(\theta)$). It is therefore a straightforward matter to include the consistency condition for training, and the extra loss term can be excluded or altered according to the problem setting.

%%%%%%%%%%%%%%%%%%%% Results %%%%%%%%%%%%%%%%%%%%%%%%%%%%%%%
\section{Results}\label{sec:Results}

\subsection{Experimental Design}
\begin{figure}[!t]
    \centering
    \scalebox{.3}{%% To modify the format of this figure refer to preamble.sty
    %%%%% Control Models Definition that Includes MaxPooling  %%%%%
\begin{tikzpicture}[
    scale=1.2,
    shorten >=1pt,->,draw=black!70, node distance=\layersep,
    neuron/.style={circle,fill=black!25,minimum size=20,inner sep=0},
    edge/.style 2 args={pos={(mod(#1+#2,2)+1)*0.33}, font=\tiny},
    distro/.style 2 args={
        edge={#1}{#2}, node contents={}, minimum size=1cm, path picture={\draw[double=orange,white,ultra thick,double distance=1pt,shorten >=0pt] plot[variable=\t,domain=-1:1,samples=51] ({\t},{0.2*exp(-100*(\t-0.05*(#1-1))^2 - 3*\t*#2))});}
      },
    weight/.style 2 args={
        edge={#1}{#2}, node contents={\pgfmathparse{0.35*#1-#2*0.15}\pgfmathprintnumber[fixed]{\pgfmathresult}}, fill=white, inner sep=2pt
      }
  ]

  \begin{scope}[xshift=10cm]
    \nncontrol{control_models} % with max pooling 
  \end{scope}

\end{tikzpicture}
%\end{document}
}
    \caption{We compare the BTCNN and BTCNN with CC against three models. Each model has four hidden layers. Refer to Table \ref{tab:control_model_specifics} for further details of the layer types in each model. }
    \label{fig:control_models_arch}
\end{figure}
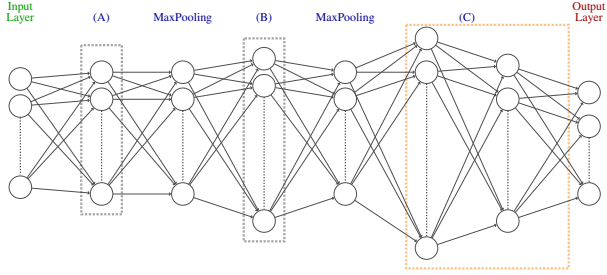

In this section, the following three models are compared against our proposed model: (1) a CNN, (2) a TCNN, and (3) a BNN. The same workflow is followed by all neural network architectures in this experiment by having two convolutional layers, each followed by max pooling and a ReLU activation function, and two fully connectedlayers, the first with ReLU activations and the second with softmax. The same hyperparameters are used by each model. Fig. \ref{fig:control_models_arch} and Table \ref{tab:control_model_specifics} provide a visualization of these models and details.

\begin{table}[!t]
    \caption{Control Model Architecture Details}
    \label{tab:control_model_specifics}
    \centering
    \begin{tabular}{|c||c|c|c|}
    \hline 
    \begin{tabular}[c]{@{}c@{}}Control \\ Model\end{tabular} & (A) & (B) & (C) \\
    \hhline{|=|=|=|=|}
    CNN & \begin{tabular}[c]{@{}c@{}}Standard \\ Convolutional \\ Layer\end{tabular} & \begin{tabular}[c]{@{}c@{}}Standard \\ Convolutional \\ Layer\end{tabular} & \begin{tabular}[c]{@{}c@{}}Standard \\ Dense Layers\end{tabular} \\
    \hline
    TCNN & \begin{tabular}[c]{@{}c@{}}Circle Filter \\ Layer\end{tabular} & \begin{tabular}[c]{@{}c@{}}Circle One \\ Layer\end{tabular} & \begin{tabular}[c]{@{}c@{}}Standard \\ Dense Layers\end{tabular} \\
    \hline
    BNN & \begin{tabular}[c]{@{}c@{}}Standard \\ Convolutional \\ Layer\end{tabular} & \begin{tabular}[c]{@{}c@{}}Standard \\ Convolutional \\ Layer\end{tabular} & \begin{tabular}[c]{@{}c@{}}Bayesian \\ Dense Layers\end{tabular} \\
    \hline
    \end{tabular}
\end{table}

For the TCNN, the first convolutional layer is a CF layer, the second is a COL, and the final two layers are dense.
The convolutional layers of the BNN are the same as that of a standard CNN. A prior of the form $p(\theta)=\mathcal{N}(\mathbf{0},\mathbf{I})$ is placed on the weights of the Bayesian fully connected layers and use a variational distribution of the form $q_\phi(\theta)=\mathcal{N}(\mu,\Sigma)$ where the covariance matrix $\Sigma$ is diagonal. This means the dense layers in the BNN will have twice as many learnable parameters compared to the dense layers in the CNN. A standard normal prior and a normal variational posterior with diagonal covariance are also used for the BTCNN and the BTCNN with a consistency condition (BTCNN with CC). Throughout the numerical results, 10 training runs are conducted for each model, and the values presented in the figures are the averages of the respective metrics across these 10 training runs.

\subsection{Model Calibration}
To evaluate the performance of the models, it is common practice to assess their calibration in addition to their predictive accuracy \cite{hands_on_BNN,primer_on_BNNs}. A well-calibrated network will show neither drastic overconfidence nor under-confidence in its predictions, but modern deep neural networks used for image classification have been shown to be poorly calibrated \cite{onCalibration}. Various methods have been proposed to address this issue, and BNNs in particular often outperform their deterministic counterparts on measures of calibration \cite{krishnan2020improving,primer_on_BNNs}. Two standard metrics for assessing neural network calibration are the expected calibration error (ECE) and maximum calibration error (MCE) \cite{Pakdaman_Naeini_Cooper_Hauskrecht_2015}. To obtain these metrics, a network's predictions are first grouped into $M$ bins $B_1, \dots, B_M$ where bin $B_m$ contains the indices $n$ of all predictions $\hat{\mathbf{p}}^{(n)}$ such that $\max_j [\hat{\mathbf{p}}^{(n)}]_{j} \in((m-1)/M,m/M]$, where $\hat{\mathbf{p}}^{(n)}$ is the prediction defined by Equation \ref{approx bma} on the $n^{th}$ input. ECE and MCE can then be computed as 
\begin{align*}
    \text{ECE} &= \sum_{m=1}^M\frac{|B_m|}{N}|\text{acc}(B_m)-\text{conf}(B_m)| \\
    \text{MCE} &= \max_{1\leq m\leq M}|\text{acc}(B_m)-\text{conf}(B_m)|
\end{align*}
where
\begin{align*}
    \text{acc}(B_m) &= \frac{1}{|B_m|}\sum_{n\in B_m}\mathbf{1}(\hat{y}_n=y_m) \\
    \text{conf}(B_m) &= \frac{1}{|B_m|}\sum_{n\in B_m}\max_j [\hat{\mathbf{p}}^{(n)}]_{j}.
\end{align*}
The accuracy of bin $B_m$, denoted by $\text{acc}(B_m)$, calculates the accuracy of the predictions in bin $B_m$ in the usual sense, where $\mathbf{1}$ is the indicator function. The confidence of bin $m$, denoted by $\text{conf}(B_m)$, calculates the average probability of the predictions in bin $B_m$. For any neural network, the relationship $0\leq\text{ECE}\leq\text{MCE}\leq1$ holds, and for a perfectly calibrated neural network, $\text{ECE}=\text{MCE}=0$.

\subsection{USPS Dataset}

\begin{figure*}[!t]
    \centering
    \includegraphics[width=0.8\textwidth]{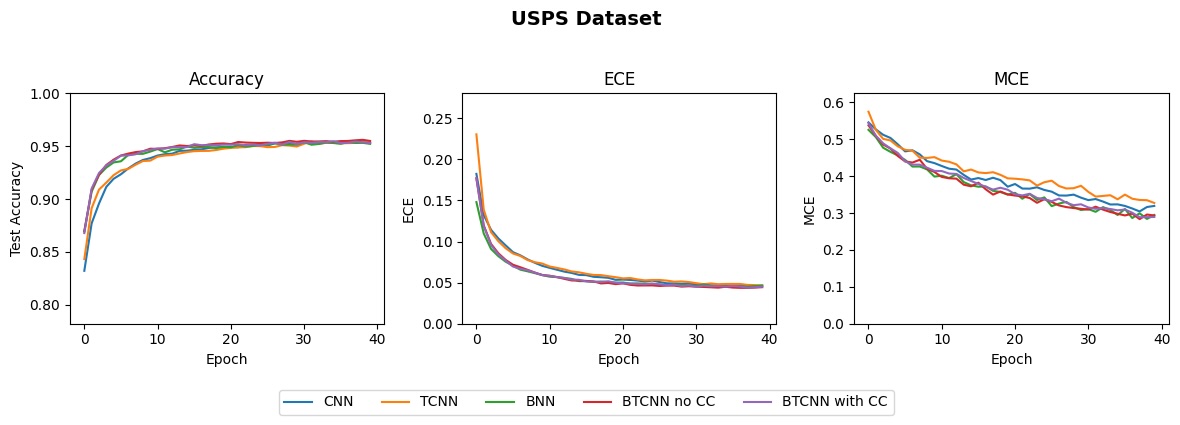}
    \caption{Comparison of the performance of CNN, BNN, TCNN, and BTCNN models both with and without a consistency condition on the USPS dataset. All of the models follow the architecture provided in Fig. \ref{fig:btcnn_layers}.}
    \label{fig:usps_40_epochs}
\end{figure*}

\begin{figure*}[!t]
    \centering
    \subfloat[]{\includegraphics[width=0.45\textwidth]{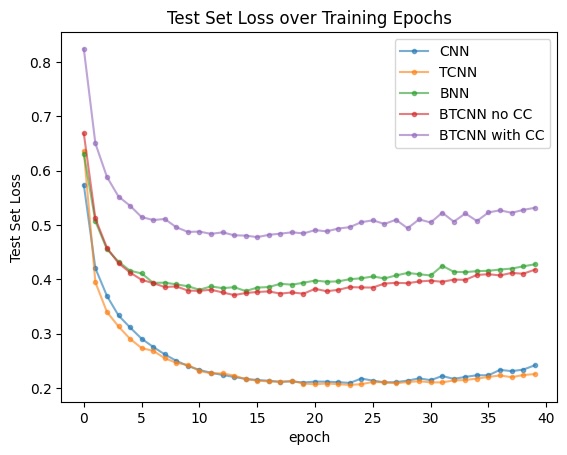}
    \label{fig:testloss_40_epochs}}
    \hfil
    \subfloat[]{\includegraphics[width=0.45\textwidth]{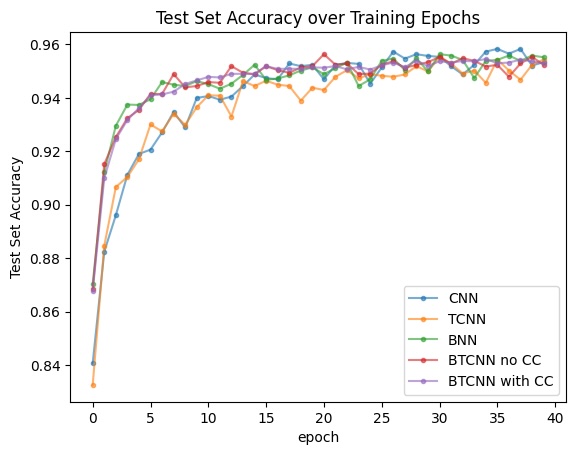}
    \label{fig:testaccurcy_40_epochs}}
    \caption{After several training epochs, the test loss eventually increases, indicating overfitting. The test accuracy for each model is converging, thus, the models are learning from the provided data. }
\end{figure*}

In this study, the USPS dataset is utilized, a freely available and popular benchmark dataset. The USPS dataset contains 9,298 images of handwritten digits from scanned pieces of mail. Each image is 16x16 pixels, greyscale, and is paired with its appropriate digit label.

From Fig. \ref{fig:usps_40_epochs}, it can be seen that a higher test accuracy is achieved by the Bayesian networks (BNN, BTCNN no CC, and BTCNN with CC) in a shorter number of epochs than by the non-Bayesian networks (CNN and TCNN). It can also be seen in the ECE and MCE plots that a lower calibration error is achieved by the Bayesian networks.

In Fig. \ref{fig:testloss_40_epochs}, it can be seen that the test loss for the Bayesian networks (BNN, BTCNN no CC, and BTCNN with CC) levels out with fewer epochs. The test loss for the Bayesian networks levels off after about 10 epochs, whereas the test loss for the non-Bayesian networks levels off after about 15 epochs. The epoch at which the test loss converges indicates that the network weights are no longer being improved and thus the network is done learning. The computational burden of neural training is reduced when the network learns faster. It should be noted that the values of the test losses are not being compared against each other, since additional terms present in the loss functions used to train the Bayesian networks are not found in the non-Bayesian networks. The curves formed by the test loss of each network are being compared. From Fig. \ref{fig:testaccurcy_40_epochs}, it can be seen that the test accuracy of the Bayesian networks is increasing much faster than that of the non-Bayesian networks, indicating that faster learning is being exhibited by the Bayesian networks.

\subsection{Data Starvation Setting}

\begin{figure*}[!t]
    \centering 
    \includegraphics[width=0.8\textwidth]{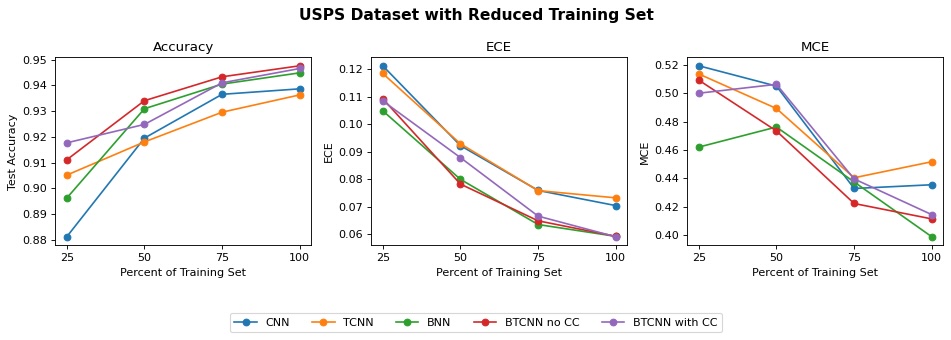}
    \caption{To summarize the results of the models' performance when trained on a reduced training set, this figure displays the respective metrics at the 10th training epoch for each subset of the USPS training set.}
    \label{fig:reduced_training_set}
\end{figure*}

To compare the performance of the networks in a data-poor setting, the networks were trained on various subsets of the USPS dataset. The subsets were randomly selected and consisted of 25\%, 50\%, 75\%, and 100\% of the original training set in the USPS dataset. In Fig. \ref{fig:reduced_training_set}, the CNN, BNN, and TCNN were generally outperformed by the BTCNNs, especially when the training set was significantly reduced (25\% of the original training set). A practical benefit is demonstrated through strong performance in circumstances where data are limited, as may occur due to challenges in sensing from faulty equipment and/or the presence of impulsive noise. In the test accuracy plots, the highest test accuracy is observed for the Bayesian networks, indicating their superior performance in data-poor scenarios. In the ECE plot, a noticeable gap is observed between the ECE of the Bayesian and non-Bayesian networks when the training dataset is reduced, suggesting that the Bayesian networks are better calibrated to the smaller training sets.  

\subsection{Gaussian Blur Applied to the USPS Dataset}
When image processing is performed on a real dataset, noise is commonly present in the data due to imperfections in the sensing hardware. Class-correlated Gaussian blurring was imposed on the USPS dataset to test our model's performance with a noisy dataset. 
To add this type of noise, a subset for each digit class was created. Then, class-correlated Gaussian blurring was applied using a Gaussian kernel where the kernel has a standard deviation that was uniformly chosen from a defined interval for each digit class.  

\begin{figure}[!t]
    \centering \includegraphics[width=0.8\linewidth]{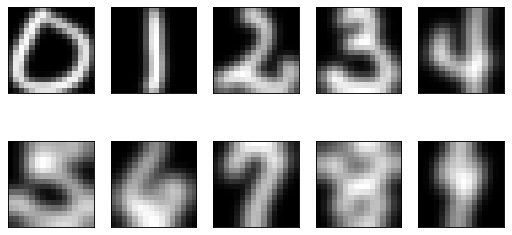}
    \caption{USPS images with class-correlated Gaussian blur}
    \label{fig:cl_corr_sample}
\end{figure}

The standard deviation intervals were defined in an increasing manner such that the ``2'' class has more noise than the ``1'' class, the ``3'' class has more noise than the ``2'' class, and so on. Refer to Fig. \ref{fig:cl_corr_sample} for a visualization of class-correlated Gaussian blurring applied to the USPS dataset. 

\begin{figure*}[!t]
    \centering 
    \includegraphics[width=0.8\textwidth]{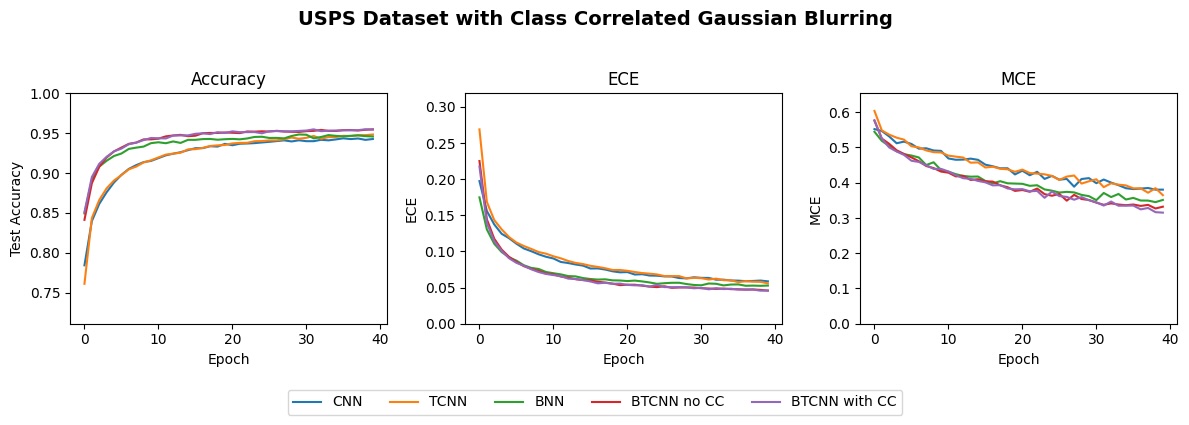}
    \caption{Comparing performance of the models on the USPS dataset with class-correlated Gaussian blur where the models train for 40 epochs. }
    \label{fig:usps_ccgblur_40epochs}
\end{figure*}

When the various models are allowed to train for 40 epochs, it can be seen, in Fig. \ref{fig:usps_ccgblur_40epochs}, that the final test accuracy is reached in fewer epochs by the BTCNN and BTCNN with CC than by the BNN and non-Bayesian networks. At about epoch 20, the training accuracy of the BTCNN and BTCNN with CC has leveled off, whereas the test accuracies of the BNN and non-Bayesian networks continue to increase into later epochs. For the ECE, a noticeable gap is observed between the Bayesian and non-Bayesian networks in the first 20 epochs, with better calibration being shown by the Bayesian networks. Eventually, the Bayesian networks are "caught up" to by the non-Bayesian networks after enough training epochs. The best performance on the MCE metric is consistently achieved by the BTCNNs, with the lowest MCE being maintained throughout the 40 training epochs.

\begin{figure*}[!t]
    \centering
    \subfloat[]{\includegraphics[width=0.45\textwidth]{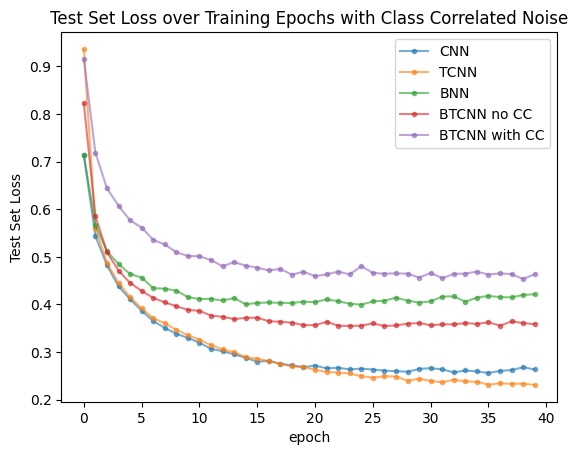}
    \label{fig:gblur_testloss_40_epochs}}
    \hfil
    \subfloat[]{\includegraphics[width=0.45\textwidth]{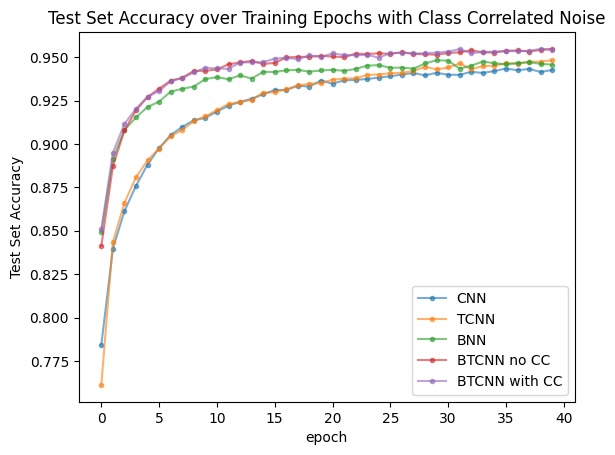}
    \label{fig:gblur_testaccurcy_40_epochs}}
    \caption{Test set loss and accuracy over 40 training epochs.}
\end{figure*}

Fig. \ref{fig:gblur_testloss_40_epochs} depicts that the Bayesian networks converged to their minimal test set losses in fewer training epochs than the non-Bayesian networks, with the Bayesian networks converging in approximately 12 epochs versus the non-Bayesian networks converging in approximately 20 epochs. It can be seen around epochs 35-40 that the test loss for most of the networks begins to increase, indicating overfitting. In Fig. \ref{fig:gblur_testaccurcy_40_epochs}, it can be seen that the BTCNNs appear to hold the highest test accuracy over the training epochs. At epochs 35-40, the test accuracy appears to decrease slightly, which is in line with the presence of overfitting at the same range of epochs concluded from Fig. \ref{fig:gblur_testloss_40_epochs}.

\begin{figure*}[!t]
    \centering 
    \includegraphics[width=0.8\textwidth]{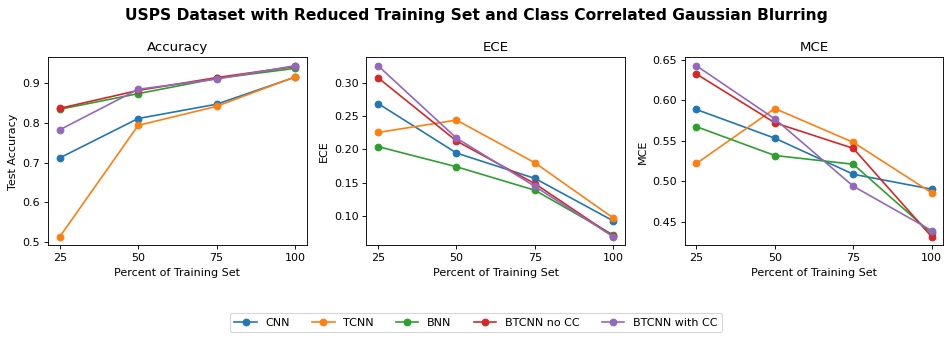}
    \caption{Model results at the final training epoch when trained with a reduced training set size and class-correlated Gaussian blurring on both the training and testing sets.}
    \label{fig:usps_ccgblur_reduced_training_set_summary}
\end{figure*}

To summarize the results of the models' performance when trained on a reduced training set with class-correlated Gaussian blur, Fig. \ref{fig:usps_ccgblur_reduced_training_set_summary} displays the respective metrics at the final training epoch. It is observed that a higher test accuracy is achieved by the Bayesian networks when the training set is reduced, and the highest test accuracy continues to be achieved as the training set increases. 
%The Bayesian networks also show the lowest ECE and MCE for the different training set sizes.

\subsection{Uncertainty Quantification}

In practice, both good accuracy and low uncertainty are desired. It is expected that the model generalizes well beyond the training set, but a tipping point is encountered when out-of-distribution data is presented. Models should indicate when they are uncertain or when their predictions should not be trusted. 
In the following experiment, the BNN, BTCNN, and BTCNN with CC models will be trained on the original USPS dataset using the original-sized training set without class-correlated Gaussian blurring. Two images of different classes will be selected from the test set, and a set of 10 images will be generated using pixel-wise convex combinations that are parameterized by weighing coefficient $\alpha$. Near the midpoint of the convex combination, the input images are heavily deviated from the learned manifold and become out-of-distribution examples.  The epistemic and total uncertainties will be approximated for each of the 10 images using each model. In this experiment, the models are challenged with out-of-distribution data to determine whether uncertainty increases when the input image lies off the learnt manifold.

\begin{figure}[!t]
    \centering
    \includegraphics[width=0.8\linewidth]{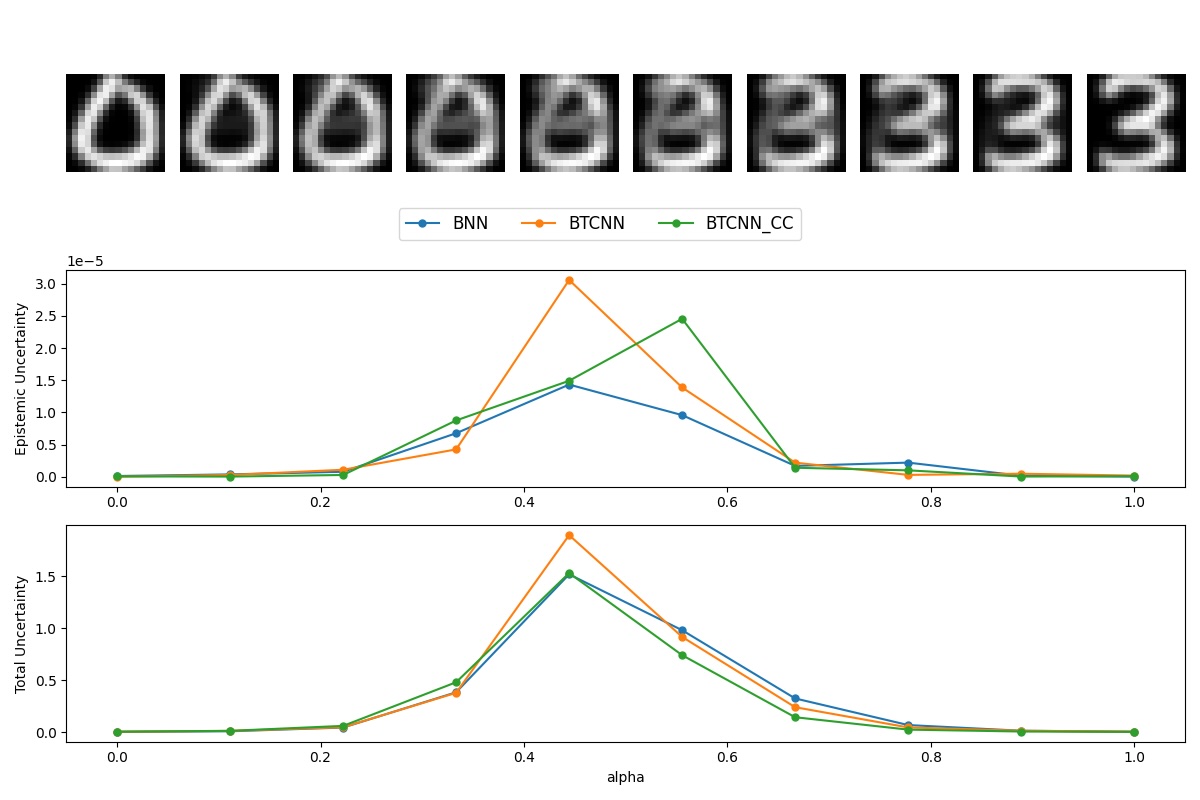}
    \caption{Epistemic uncertainty (middle) and total uncertainty (bottom) for the BNN, BTCNN, and BTCNN with CC models along a convex combination between an image of the ``0'' and ``3'' classes from the USPS test dataset where the convex combination is parameterized by alpha.}
    \label{fig:uncert_03}
\end{figure}

In Fig. \ref{fig:uncert_03}, it is observed that both the epistemic and total uncertainty are high near the midpoint of the convex combination (further from the learned manifold) and low near the endpoints of the convex combination (closer to the learned manifold). The highest epistemic uncertainty values were achieved near the midpoint of the convex combination by the BTCNN and BTCNN with CC. The BTCNN and BTCNN with CC strike a balance between showing sensitivity to out-of-distribution data, especially with respect to the epistemic uncertainty, and having low uncertainty for images near the learned manifold ($\alpha$ values near 0 or 1).

\begin{figure}[!t]
    \centering
    \includegraphics[width=0.8\linewidth]{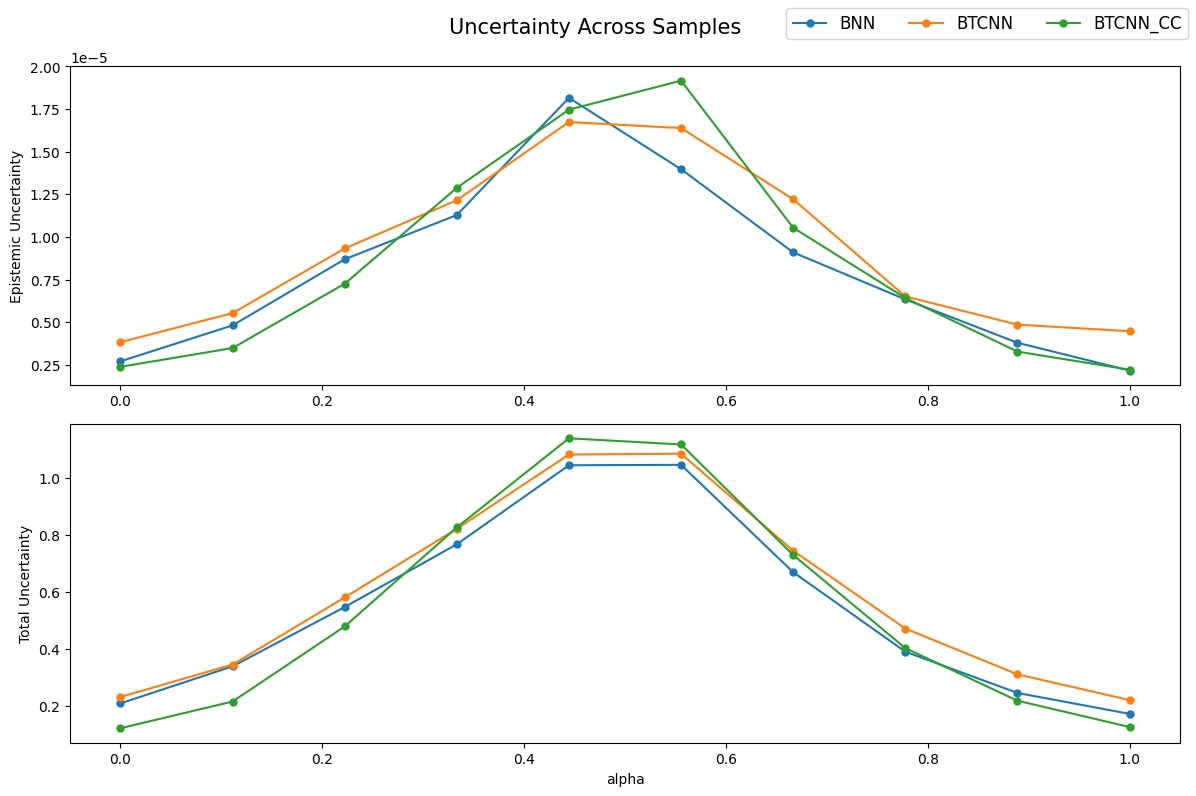}
    \caption{Average epistemic uncertainty (top) and average total uncertainty (bottom) for each alpha value across all digit pairs ($45$ in total).}
    \label{fig:avg_uncert}
\end{figure}

To understand the trends of the epistemic and total uncertainties across all unique digit pairs, the uncertainty value for each alpha is averaged for each model in Fig. \ref{fig:avg_uncert}. In general, the anticipated behavior of increased uncertainty is observed near the midpoint of the convex combination, where the input images are furthest from the learned manifold. This indicates that the BNN, BTCNN, and BTCNN with CC are generally certain about in-distribution examples and generally uncertain about out-of-distribution examples. In both the epistemic and total uncertainty, the lowest uncertainty near the endpoints of the convex combination (closer to the learned manifold) and the highest uncertainty is observed near the midpoint of the convex combination (further from the learned manifold) both achieved by the BTCNN with CC. The BTCNN with CC achieves a balance between certainty with in-distribution examples and appropriate uncertainty to out-of-distribution examples.

%%%%%%%%%%%%%%%%%%%% Discussion %%%%%%%%%%%%%%%%%%%%%%%%%%
\section{Discussion}\label{sec:Discussion}
We have introduced a novel Bayesian topological CNN learning architecture, which utilizes information from manifolds while reducing calibration error and improving uncertainty quantification by placing prior distributions on network parameters and learning the corresponding posteriors. We illustrate how the prior distributions can be modified to promote desired behavior in the network via the inclusion of a consistency condition and then evaluate our model on a benchmark image classification dataset to demonstrate its superiority over conventional methods, particularly in cases involving limited or noisy data.

There are several potential refinements and extensions of our method that could be explored in future research. During our testing, we found that introducing stochasticity into the topological layers actually degraded their performance, so we did not place prior distributions on the parameters of these layers in our experiments. Placing priors only on the final dense layers improved the network's calibration, but more work can be done to investigate why this occurs and whether there are prior distributions that do not hinder the topological layers. In addition, our method relies on discretizations of a single topological manifold, the unit circle $S^1$, but the optimal manifold could depend on the particular task and architecture being used. Future investigations could analyze the effectiveness of alternative manifolds or even learning the manifold as part of the training process. Moreover, the incorporation of topological information need not be restricted to the network's convolutional layers. Alternative modifications of the network priors via consistency conditions that take into account topological features of the training set could further improve the performance of the BTCNN and allow our framework to be extended to other non-convolutional neural network architectures.

%%%%%%%%%%%%%%%%%%%% Acknowledgement %%%%%%%%%%%%%%%%%%%%%%%
\section*{Acknowledgements}
This work has been partially supported by the Army Research Laboratory Cooperative Agreement No. W911NF2120186, STRONG ARL CA No. W911NF-22-2-0139, and The University of Tennessee Materials Research Science \& Engineering Center – The Center for Advanced Materials and Manufacturing – NSF DMR No. 2309083.
%%%%%%%%%%%%%%%%%%%% Bibliography %%%%%%%%%%%%%%%%%%%%%%%%%%
\addcontentsline{toc}{section}{Bibliography}
\nocite{*}
\printbibliography

\end{document}